\documentclass[10pt,twocolumn,letterpaper]{article}

\usepackage[pagenumbers]{cvpr} 

\usepackage{graphicx}
\usepackage{amsmath}
\usepackage{amssymb}
\usepackage{booktabs}
\usepackage{multirow}

\usepackage{caption}

\usepackage[pagebackref,breaklinks,colorlinks]{hyperref}
\usepackage{floatrow}

\def\eg{\emph{e.g.}} 
\def\ie{\emph{i.e.}} \def\Ie{\emph{I.e.}}

\usepackage[capitalize]{cleveref}
\crefname{section}{Sec.}{Secs.}
\Crefname{section}{Section}{Sections}
\Crefname{table}{Table}{Tables}
\crefname{table}{Tab.}{Tabs.}

\def\cvprPaperID{716} 
\def\confName{CVPR}
\def\confYear{2022}

\begin{document}

\title{AziNorm: Exploiting the Radial Symmetry of Point Cloud \\for Azimuth-Normalized 3D Perception}

\author{
{Shaoyu Chen} \textsuperscript{1, 2} \quad
 {Xinggang Wang} \textsuperscript{1}$^\dag$ \quad {Tianheng Cheng} \textsuperscript{1, 2} \quad {Wenqiang Zhang} \textsuperscript{1} \\
{Qian Zhang} \textsuperscript{2} \quad {Chang Huang} \textsuperscript{2} \quad  {Wenyu Liu} \textsuperscript{1}
\\
\textsuperscript{1} School of EIC, Huazhong University of Science \& Technology \quad \textsuperscript{2} Horizon Robotics \\
{\tt\small \{shaoyuchen,xgwang,thch,wq\_zhang,liuwy\}@hust.edu.cn \quad \{qian01.zhang, chang.huang\}@horizon.ai }}

\maketitle

\begin{abstract}
Studying the inherent symmetry of data is of great importance in machine learning. Point cloud, the most important data format for 3D environmental perception, is naturally endowed with strong radial symmetry. In this work, we exploit this radial symmetry via a divide-and-conquer strategy to boost 3D perception performance and ease optimization. We propose Azimuth Normalization (AziNorm), which normalizes the point clouds along the radial direction and eliminates the variability brought by the difference of azimuth. AziNorm can be flexibly incorporated into most LiDAR-based perception methods. To validate its effectiveness and generalization ability, we apply AziNorm in both object detection and semantic segmentation. For detection, we integrate AziNorm into two representative detection methods, the one-stage SECOND detector and the state-of-the-art two-stage PV-RCNN detector. Experiments on Waymo Open Dataset demonstrate that AziNorm improves SECOND and PV-RCNN by $7.03$ mAPH and $3.01$ mAPH respectively. For segmentation, we integrate AziNorm into KPConv. On SemanticKitti dataset, AziNorm improves KPConv by $1.6/1.1$ mIoU on val/test set. Besides, AziNorm remarkably improves data efficiency and accelerates convergence, reducing the requirement of data amounts or training epochs by an order of magnitude. SECOND w/ AziNorm can significantly outperform fully trained vanilla SECOND, even trained with only $10\%$ data or $10\%$ epochs. Code and models
are available at \url{https://github.com/hustvl/AziNorm}.
\end{abstract}

\begin{figure}[]
    \centering
    \includegraphics[width=0.88\linewidth]{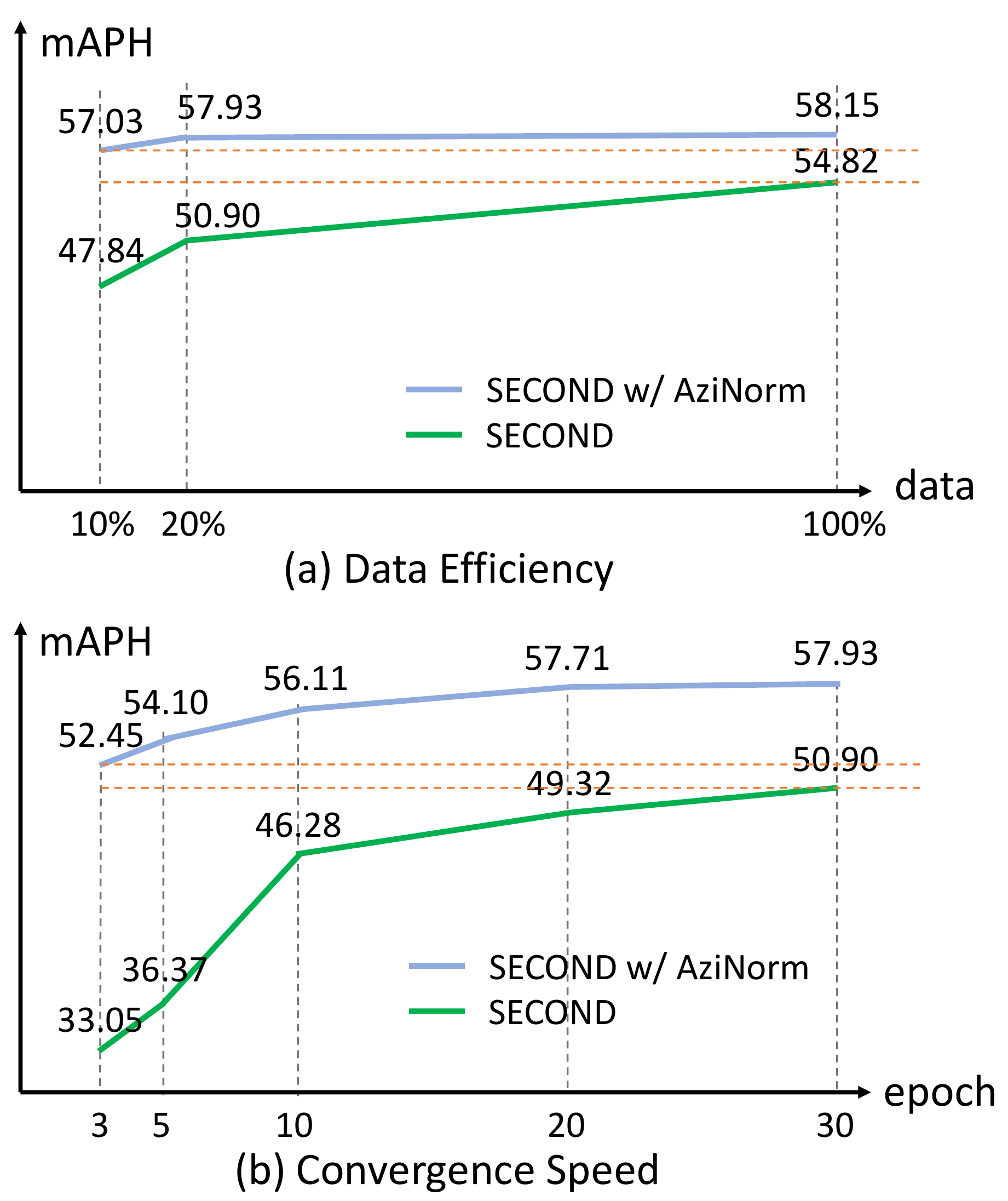}
    \caption{\textbf{AziNorm remarkably improves data efficiency (a) and accelerates convergence (b).}  With data amounts and training epochs reduced, the performance of SECOND drops dramatically, but SECOND w/ AziNorm still achieves comparable performance. Even trained with only  $10\%$ data or $10\%$ epochs, SECOND w/ AziNorm obviously outperforms fully trained vanilla SECOND.}
    \label{fig:convergence}
\end{figure}

\section{Introduction}
\let\thefootnote\relax\footnotetext{$^\dag$Xinggang Wang is the corresponding author.}

Environmental perception based on 3D LiDAR point clouds is a fundamental and indispensable ability for autopilot system aiming at high robustness and security. Accurate perception results are the basis of credible motion planning and control, avoiding traffic accidents.
And the passed several years have witnessed tremendous performance improvements of LiDAR-based perception.
Previous works~\cite{qi2017frustum,shi2019pointrcnn,std2019yang,qi2019deep,yang20203dssd,zhou2018voxelnet,yang2018pixor,ye2020hvnet,second,shi2020part} mostly focus on designing network architectures to effectively and efficiently extract semantic information from the point cloud data.

Different from the exhaustively studied network architecture design, the inherent property of LiDAR point cloud has not been explored and exploited yet.
LiDAR emits laser rays that travel outwards in all azimuths \footnote{In 3D perception, azimuth is the horizontal direction expressed as the angular distance between the direction of observer's heading and the direction of the observed location.} isotropically. Laser rays bounce off surrounding objects, capturing object surface characteristics, and then return to the LiDAR.
Both incident rays and reflected rays are along the radial direction. And the generated point clouds are radially symmetrical in a broad sense (Fig.~\ref{fig:pipeline}).

From the perspective of machine learning, the inherent symmetry of data can serve as a powerful inductive bias for reducing the variability of data and simplifying the recognition system.
In this paper, we propose \textbf{Azimuth Normalization} (\textbf{AziNorm}) to exploit the radial symmetry of point cloud.
We split the whole LiDAR scene into individual patches and then normalize the patch's sub point clouds along the radial direction. By normalization, we eliminate the data variability brought by the difference of azimuth and makes perception on normalized point clouds much easier.

AziNorm can be flexibly incorporated into most LiDAR-based perception methods without modifying any implementation detail and significantly boost the performance. 
To validate its effectiveness and generalization ability, we apply AziNorm in both object detection and semantic segmentation.
For detection, we integrate AziNorm into two representative detection methods,
the one-stage SECOND~\cite{second} detector and the state-of-the-art two-stage PV-RCNN~\cite{pvrcnn} detector. Experiments on large-scale Waymo Open Dataset~\cite{waymo} demonstrate that AziNorm improves SECOND and PV-RCNN by $7.03$ mAPH and $3.01$ mAPH respectively.
For segmentation, we integrate AziNorm into KPConv~\cite{KPConv}. On SemanticKitti~\cite{SemanticKITTI} dataset, AziNorm improves KPConv by $1.6/1.1$ mIoU on val/test set.

More importantly, AziNorm remarkably improves data efficiency and accelerates convergence. Even trained with only $10\%$ data or $10\%$ epochs, SECOND w/ AziNorm significantly outperforms fully trained vanilla SECOND. 
AziNorm can reduce the requirement of training epochs or data amounts by an order of magnitude, lowering the cost of data acquisition and labeling, which is of great practical value, especially for the data-driven autopilot system.

The main contributions of this paper can be
summarized as follows:
\begin{itemize}
\item We propose a novel normalization method termed as AziNorm, which exploits the inherent radial symmetry of point clouds to reduce the data variability.
\item AziNorm can be easily integrated into most LiDAR-based perception methods and significantly boost the performance. We validate the effectiveness and generalization ability of AziNorm in two perception tasks (object detection and semantic segmentation), based on three methods (SECOND~\cite{second}, PV-RCNN~\cite{pvrcnn} and KPConv~\cite{KPConv}), and on two datasets (Waymo Open Dataset~\cite{waymo} and SemanticKitti~\cite{SemanticKITTI}).
\item 
AziNorm can reduce the requirement of data amounts or training epochs by an order of magnitude, lowering the cost of data acquisition and labeling, which is of great practical value, especially for the data-driven autopilot system.
\end{itemize}

\section{Related Work}

\subsection{Representation Learning on Point Clouds}
Representation learning on point clouds has made tremendous progress in recent years \cite{qi2017pointnet,qi2017pointnet++,zhou2018voxelnet,wang2019dynamic,huang2018recurrent,zhao2019pointweb,li2018pointcnn,su2018splatnet,wu2019pointconv,jaritz2019multi,jiang2019hierarchical,thomas2019kpconv,choy20194d,liu2020closer,xu2020grid}. 
The popular PointNet series~\cite{qi2017pointnet,qi2017pointnet++} proposes to directly learn point-wise features from the raw point clouds, where set abstraction operation enables flexible receptive fields by setting different searching radii.
3D sparse convolution~\cite{Submanifold,3DSegSubmanifold} is adopted in~\cite{second,shi2020part} to effectively learn sparse grid-wise features from the point clouds. 
\cite{liu2019point} combines both grid-based CNN and point-based shared-parameter multi-layer perceptron (MLP) network for point cloud feature learning. 

Because of the high complexity and irregularity of point cloud, large amounts of data and long training time are required for learning the representation of point cloud. 
AziNorm normalizes point clouds and unifies symmetric patterns. It makes representation learning on point clouds much easier and improves both data efficiency and convergence speed.

\subsection{3D Perception on Point Clouds}
3D perception on point clouds aims to extract semantic information from the unordered and sparse point cloud data.
Existing methods can be divided into three categories, \ie, point-based, grid-based and range-based methods, according to the formats of point cloud they work on.

\noindent\textbf{Grid-based Perception Methods} \ 
The irregular data format of point cloud brings great challenges for 3D perception.
Grid-based methods generally project the point clouds to regular bird-eye view grids~\cite{Chen2017CVPR,yang2018pixor} or 3D voxels \cite{zhou2018voxelnet,Chen2019fastpointrcnn} for processing point clouds with 2D/3D CNN.
The pioneer work MV3D~\cite{Chen2017CVPR} projects the point clouds to 2D bird-eye view grids and the following works~\cite{ku2018joint,Liang2018ECCV,Liang2019CVPR,vora2020pointpainting,yoo20203d,huang2020epnet} develop better strategies for multi-sensor fusion. \cite{yang2018pixor,Yang2018CoRL,lang2018pointpillars} introduce more efficient frameworks with bird-eye view representation.
Some other works~\cite{song2016deep,zhou2018voxelnet} divide the point clouds into 3D voxels to be processed by 3D CNN. 
PV-RCNN~\cite{pvrcnn} incorporates both 3D voxel CNN and PointNet-based set abstraction for learning discriminative point cloud features.
~\cite{polarnet,cylindrical} project LiDAR points into polar grids to adapt to the long-tailed distribution of LiDAR points. They help to eliminate the azimuth variability, but their unevenly partitioning the 3D space introduces the problem of scale variance. \Ie, in polar grid representation, object's size varies with its distance to LiDAR. Our AziNorm eliminates the azimuth variability without leading to scale variance.

\noindent\textbf{Point-based Perception Methods} \ 
Point-based perception methods directly operate on the original format of point clouds.
F-PointNet~\cite{qi2017frustum} first proposes to apply PointNet~\cite{qi2017pointnet,qi2017pointnet++} for 3D detection from the cropped point clouds based on the 2D image bounding boxes. PointRCNN~\cite{shi2019pointrcnn} generates 3D proposals directly from  point clouds and refines proposals with a RCNN stage. And the following work STD~\cite{std2019yang} proposes the sparse to dense strategy for better proposal refinement. \cite{qi2019deep} proposes the hough voting strategy for better object feature grouping.
KPConv~\cite{KPConv} introduces a novel spatial kernel-based point convolution.

\noindent\textbf{Range-based Perception Methods} \ 
Range-based perception methods  operate on the range image.
LaserNet~\cite{lasernet} predicts a multimodal distribution for each point to generate the final prediction. RCD~\cite{RCD} learns a dynamic dilation for scale variation and soft range gating for the “boundary blur” issue. RangeDet~\cite{rangedet}  fixes issues of scale variance and inconsistency between 2D and 3D coordinates.
RSN~\cite{rsn} predicts foreground points from range images and applies sparse convolutions on the selected foreground points to detect objects.

AziNorm is with high generalization ability. It can be integrated with grid-based,  point-based and some hybrid methods.
And existing 3D perception methods treat the LiDAR scene as a whole for processing.
AziNorm introduces the divide-and-conquer strategy, which splits the scene into individual patches. With AziNorm, point clouds can be processed in a flexible manner.

\subsection{Normalization}
Broadly speaking, normalization is to reduce the variability of data. 
Normalizing the data benefits training a lot~\cite{LeCun_backprop}.
Batch Normalization~\cite{bn} is a representative normalization technique, which eases optimization and enables deep networks to converge.

In 3D domain, RoI pooling can be considered as an object-level normalization. Some two-stage detectors~\cite{shi2019pointrcnn, shi2020part, pvrcnn} adopt RoI pooling in the second stage, transforming points to the canonical coordinate system, whose XYZ axes are parallel to the predicted 3D bounding boxes. RoI pooling just unifies objects. It does not exploit the property of point cloud and is only applied in object detection.
Differently, AziNorm focuses on the inherent radial symmetry of LiDAR point cloud and leverages it for normalization. And AziNorm is a general normalization method, not only for object detection, but also for other LiDAR-based perception tasks. Besides, as other normalization methods do, AziNorm significantly 
improves optimization.

\begin{figure*}[]
    \centering
    \includegraphics[width=\linewidth]{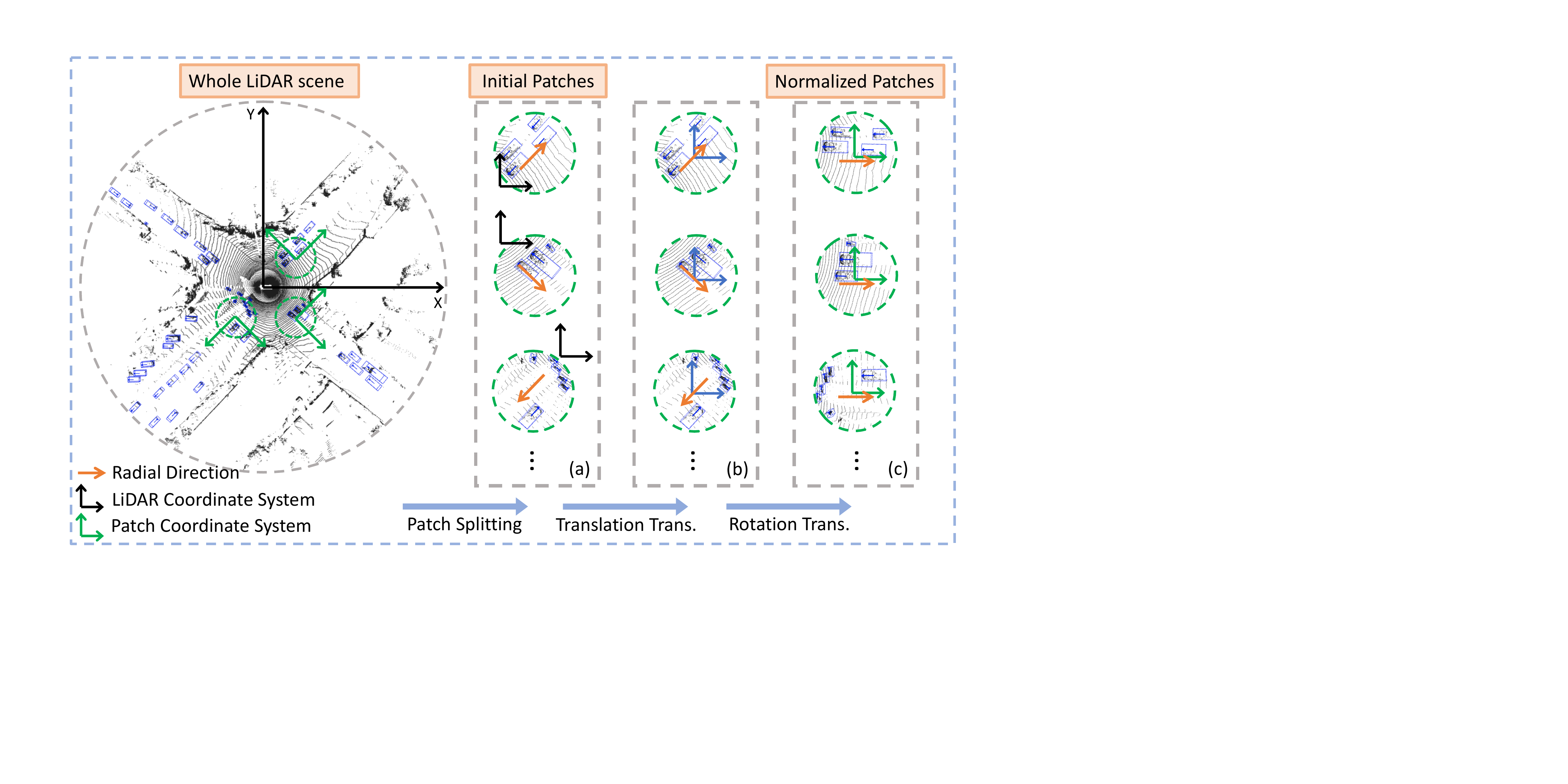}
    \caption{\textbf{Illustration of Azimuth Normalization.} The whole LiDAR scene is split into overlapped circular patches. Within each patch, we transform the coordinates of the sub point clouds from the LiDAR coordinate system to the specific patch coordinate system (translation and rotation transformation sequentially). With AziNorm, the radial directions of all patches are unified and the variability of azimuth is eliminated. Perception based on normalized sub point clouds is significantly simplified. For clarity, only three typical patches are visualized.
    In (a), (b), and (c), the coordinates of point clouds  are relative to the LiDAR coordinate system, the temporary coordinate system colored in blue, and the patch coordinate system, respectively.}
    \label{fig:pipeline}
\end{figure*}

\section{Method} 
AziNorm functions as a point cloud normalization method to eliminate the variability of azimuth, based on the inherent radial symmetry.
We adopt a \textbf{divide-and-conquer} strategy to normalize the point clouds along the radial direction.
\Ie, we split the whole LiDAR scene into patches and normalize in the patch level. We then conduct patch-wise perception and  merge patch-wise predictions (see Fig.~\ref{fig:pipeline}). The details are as follows.

\noindent\textbf{Patch Splitting} \   
Firstly, we split the whole LiDAR scene into overlapped circular patches $\{P_i\}$ with stride $d$ and radius $r$. The layout of patch is optional (discussed in Sec.~\ref{sec:abla}) and circular patches are chosen for method description. Each patch corresponds to sub point clouds denoted as $P_i =\{p^i_j,j \in [1,N_i]\}$, where ${N_i}$ is the number of points in the $i$-th patch.
The center of patch $P_i$ is denoted as $c_i$.
And we denote the azimuth of patch $P_i$ as $\theta_i$, which is the polar angle of the patch center $c_i$ against positive X-axis of the LiDAR coordinate system.

For the whole LiDAR scene, the azimuths vary from $0^\circ$ to $360^\circ$.
While for a patch, because of the limited spatial range, the azimuths of all points vary little and the difference is negligible.
Thus, all the points in a patch can be treated as a whole for normalization.

\noindent\textbf{Patch Selection} \ 
In the scenario of autopilot, the LiDAR scene is of large range and high sparsity (see Fig.~\ref{fig:visual}). 
Taking Waymo~\cite{waymo} dataset for example, in its extremely large scene of $150m \times 150m$,  about $80\%$ of areas are void or only contain few ground points.

Existing 3D perception methods have to regard the complete point clouds as a whole for processing. With AziNorm, we can process point clouds in a lower patch level and in a flexible manner.
And the sparsity of LiDAR scene can be leveraged to achieve higher efficiency.

We adopt two patch selection strategies, \textbf{negative filtering} and \textbf{positive sampling}, for more efficient training and inference.
Specifically, in both training and inference, we directly filter out patches only with negligible points, avoiding time consumption in background regions.
Besides, for object detection, in the training phase, we keep all foreground patches that contain ground truth objects and only sample a few background patches,
balancing the ratio between positive and negative samples.

\noindent\textbf{Patch Normalization} \ 
After patch splitting and selection, patch-specific transformation is conducted to normalize among different patches.

For every patch, we build a patch coordinate system, whose positive X-axis is along the radial direction (see Fig.~\ref{fig:pipeline}).
Within patch $i$, we transform points from the LiDAR coordinate system to the patch coordinate system with the extrinsic matrix $R_{\theta_i}$ and $T_{c_i}$,
\begin{equation}
    \bar{p}^i_j  =  R_{\theta_i}  \cdot (p^i_j +  T_{c_i}).
\end{equation}
$R_{\theta_i}$ and $T_{c_i}$ are the rotation matrix determined by $\theta_i$  and translation matrix determined by  $c_i$ respectively. $\bar{p}^i_j$ is the transformed coordinate of $p^i_j$. $\bar{p}^i_j$ and $p^i_j$ are relative to the patch coordinate system and LiDAR coordinate system respectively.
After patch-specific transformation, the radial directions of all patches are unified (see Fig.~\ref{fig:pipeline}(c)). 
The variability of azimuth is almost eliminated
and the point clouds are simplified.

\noindent\textbf{Patch-wise Perception} \ The next step is applying perception algorithms on normalized patches.
Patch's sub point clouds keep the same data format and geometric information with the original point clouds.
Thus, perception on patches is the same with perception on the whole LiDAR scene.
Most off-the-shelf LiDAR-based perception methods, \eg, both object detection and semantic segmentation, both point-based~\cite{qi2017frustum,shi2019pointrcnn,std2019yang,qi2019deep,yang20203dssd} and grid-based~\cite{zhou2018voxelnet,yang2018pixor,ye2020hvnet,second,shi2020part} ones, are applicable for patch-wise perception.

Specifically, we treat every patch as  independent point clouds and concatenate all patches in the batch dimension. 
Then a chosen perception method is adopted for both training and inference.
It's worth noting that we need not make any special modification for adapting to different perception methods. All the hyper-parameters (including voxel size, batch size, learning rate, \etc.) are kept the same with the off-the-shelf configurations.

\noindent\textbf{Inverse Normalization} \ For object detection, with the detection on patches finished, we get detected 3D bounding boxes, represented in different patch coordinate systems. We then conduct inverse normalization to convert the patch detection results $\{\bar{B}_i\}$ back to the original LiDAR coordinate system representation $\{B_i\}$, which is formulated as,
\begin{equation}
    B_i  =   R_{\theta_i}^{-1} \cdot \bar{B}_i -  T_{c_i}.
\end{equation}

For semantic segmentation, after applying segmentation methods on patches, we get predicted labels of every point of every patch.
We then map the predicted results from patches to the original point clouds.

\noindent\textbf{Patch Merging} \ 
Patches are overlapped and the predictions are duplicated. We merge patch-wise predictions to generate the final predictions. 

For object detection, through strict non-maximum suppression (NMS), duplicated bounding box predictions are filtered out.
For semantic segmentation, one point is included by several patches. For each point, we just average the predictions of different patches.

\begin{table*}
    \centering
    \small
    {
	\begin{tabular}{l  l  cc  cc  cc  cc}
                \toprule
				\multirow{2}{*}{Difficulty} &
                \multirow{2}{*}{Method} & 
				\multicolumn{2}{c}{~~ All} &  
				\multicolumn{2}{c}{~~ Vehicle} &  
				\multicolumn{2}{c}{~~Pedestrian} & 
				\multicolumn{2}{c}{~Cyclist} \\
				& &mAP & mAPH & AP & APH & AP & APH & AP & APH \\
				\midrule
			    \multirow{6}{*}{LEVEL 1} 
				& SECOND~\cite{second} & 60.57 & 56.49 & 68.15 & 67.56 & 58.20 & 47.94 & 55.35 & 53.98 \\
				& SECOND w/ AziNorm & 67.32 & 63.40	& 70.73 & 70.24 & 67.39 & 57.23 & 63.84 & 62.74\\
				& Improvement & +6.75 & +6.91 & +2.58 & +2.68 & +9.19 & +9.29 & +8.49 & +8.76 \\
				\cmidrule{2-10}
	            & PV-RCNN~\cite{pvrcnn} & 69.74 & 65.23  & 74.27 & 73.58 & 70.39 & 59.10 & 64.57 & 63.00 \\
                & PV-RCNN w/ AziNorm  & 72.64 & 68.32 & 75.17 & 74.64 & 75.05 & 63.92 & 67.69 & 66.41  \\
                & Improvement & +2.90 & +3.09 & +0.90 & +1.06 & +4.66 & +4.82 & +3.12 & +3.41 \\
   				\midrule
			    \multirow{6}{*}{LEVEL 2} 
				& SECOND~\cite{second}  & 54.48 & 50.90 & 59.61 & 59.09 & 50.22 & 41.32 & 53.62 & 52.28 \\
				& SECOND w/ AziNorm & 61.51 & 57.93 & 63.03 & 62.56  & 59.73 & 50.55 & 61.76 & 60.69\\
				& Improvement & +7.03 & +7.03 & +3.42 & +3.47 & +9.51 & +9.23 & +8.14 & +8.41\\ 
				\cmidrule{2-10}
	
	            & PV-RCNN~\cite{pvrcnn} & 63.10 & 59.04 & 65.30 & 64.68  & 61.33 & 51.31 & 62.67 & 61.14 \\
                & PV-RCNN w/ AziNorm & 65.95 & 62.05 & 66.27 & 65.80 & 65.93 & 55.97 & 65.65 & 64.39\\
                & Improvement & +2.85 & +3.01 & +0.97 & +1.12 & +4.60 & +4.66 & +2.98 & +3.25\\
				\bottomrule
			\end{tabular}}
	\caption{\textbf{Performance comparisons of object detection on the Waymo~\cite{waymo} val set.} All experiments are conducted with the same configuration based on the official codebase OpenPCDet~\cite{openpcdet2020}. AziNorm-based detectors significantly outperform their baselines in all metrics of all difficulty levels.}
	\label{tab:abla_ap}
\end{table*}

\begin{table}
    \small
    \centering
    \renewcommand{\tabcolsep}{3.5pt}
			\begin{tabular}{l  c c c}
                \toprule
				Method& Rec.@0.3 & Rec.@0.5 &Rec.@0.7 \\
				\midrule
			    SECOND~\cite{second} & 80.13 & 70.37 & 44.60 \\
				SECOND w/ AziNorm & 81.50 & 72.94 & 47.67 \\
				Improvement & +1.37 & +2.57 & +3.07\\

				\midrule
			    PV-RCNN~\cite{pvrcnn} & 84.86 & 78.12 & 53.92 \\
				PV-RCNN w/ AziNorm & 88.90 & 82.53 & 58.61 \\
				Improvement & +4.04 & +4.41 & +4.69\\
				\bottomrule
			\end{tabular}
	\caption{\textbf{Recall comparisons on the Waymo~\cite{waymo} val set under IoU thresholds of 0.3, 0.5 and 0.7.} AziNorm-based detectors achieve much higher recall rates than their baselines under all IoU thresholds.}
	\label{tab:abla_recall}
\end{table}

\section{Experiments} 

\subsection{AziNorm-based Object Detection}

\noindent\textbf{Experimental Settings} \ 
AziNorm can be flexibly incorporated into most point cloud based object detection methods. 
For validating its effectiveness and generalization ability, we integrate AziNorm into two representative  detection methods. One is the one-stage detector SECOND~\cite{second} with a compact and efficient pipeline,
and the other is the two-stage PV-RCNN~\cite{pvrcnn}, which is the state-of-the-art 3D detection method.
For fair comparison, we keep all the hyper-parameters the same with the off-the-shelf configurations provided by OpenPCDet~\cite{openpcdet2020} and do not modify any implementation detail of the detector.
As for patch splitting, unless otherwise specified, the patch's radius $r$ is $9.6m$ and the stride $d$ between two adjacent patches is $6.4m$. 
Experiments are conducted with $8$ RTX 3090 GPUs.

\noindent\textbf{Dataset and Metrics} \ 
We conduct experiments on Waymo Open Dataset~\cite{waymo}, which is the largest public dataset for autonomous driving to date. 
There are totally $798$ training sequences with around $160k$ LiDAR frames, and $202$ validation sequences with $40k$ LiDAR frames. 
Using the entire training set requires a lot of computing resources and training time. Following \cite{pvrcnn}, to efficiently conduct experiments, we uniformly sample $20\%$ frames (about $32k$ frames) for training (except for the experiments of data efficiency).
We report  mean average precision (mAP) and  mean average precision weighted by heading (mAPH) on two difficulty levels (LEVEL 1 and LEVEL 2).
The 3D IoU threshold is set as $0.7$ for vehicle detection and $0.5$ for pedestrian/cyclist detection.

\noindent\textbf{SECOND w/ AziNorm} \ As shown in Tab.~\ref{tab:abla_ap}, compared with vanilla SECOND, SECOND w/ AziNorm achieves remarkably better mAP/mAPH on all difficulty levels for the detection of all three categories.
And AziNorm improves the recall under the IoU threshold of $0.3$, $0.5$ and $0.7$ by $1.37$, $2.57$ and $3.07$ respectively (Tab.~\ref{tab:abla_recall}).
Notably, the performance of pedestrian and cyclist is improved more.  We can observe a gain of $9.23$ and $8.41$ APH of LEVEL 2 difficulty for pedestrian and cyclist respectively.
Pedestrian and cyclist occupy less space in real world and are with fewer LiDAR points on their surfaces. In 3D detection, these two classes are more challenging.
AziNorm unifies the symmetric point cloud patterns and highly reduces the learning difficulty of hard cases with few LiDAR points.
Thus, the detection of pedestrian and cyclist benefits more from AziNorm.

\noindent\textbf{PV-RCNN w/ AziNorm} \ 
As shown Tab.~\ref{tab:abla_ap}, when incorporated into the state-of-the-art detector PV-RCNN with less room for improvement, AziNorm still brings significant gain. PV-RCNN w/ AziNorm outperforms vanilla PV-RCNN in all metrics.  As for the mAPH of LEVEL 2 difficulty of all classes, AziNorm improves PV-RCNN by $3.01$ mAPH. 
And AziNorm improves the recall under the IoU threshold of $0.3$, $0.5$ and $0.7$ by $4.04$, $4.41$ and $4.69$ respectively (Tab.~\ref{tab:abla_recall}).

\begin{table*}
	\begin{center}
		\resizebox{1\linewidth}{!}
		{
		\setlength\tabcolsep{2pt}
		\begin{tabular}{l cccccccccc cccccccccc}

        \toprule
        Method&mIoU&car&bcycle.&mcycle.&truck&o-veh.&person&bclist.&mclist.&road&parking&side.&o-gro.&bui.&fence&vege.&trunk&terrain&pole&tra.\\
		\midrule
		KPConv  & 61.3 & 95.5 & 39.1 & 62.6 & 61.0 & 50.7 & 72.1 & 91.5 & 0.00 & 91.1 & 30.2 & 80.3 & 1.1 & 89.8 & 70.1 & 88.5 & 68.7 & 74.0 & 65.7 & 32.9\\
        KPConv w/ AziNorm& 62.9 & 95.9 & 42.6 & 66.3 & 67.0 & 55.1 & 73.0 & 90.3 & 0.00 & 91.4 & 33.3 & 80.5 & 2.7 & 90.1 & 70.7 & 88.9 & 70.7 & 74.6 & 66.7 & 36.1\\
        Improvement&+1.6&+0.4&+3.5&+3.7&+6.0&+4.4&+0.9&-1.2&+0.0&+0.3&+3.1&+0.2&+1.6&+0.3&+0.6&+0.4&+2.0&+0.6&+1.0&+3.2\\
        \bottomrule
		\end{tabular}
		}
	\end{center}
	\caption{\textbf{Performance comparisons of semantic segmentation on the SemanticKitti~\cite{SemanticKITTI} \textbf{val} set}. All experiments are conducted with the same configuration based on the official code of KPConv~\cite{KPConv}. AziNorm-based KPConv significantly outperforms vanilla KPConv.}
	\label{tab:semantickitti_val}
\end{table*}

\begin{table*}
    \centering
	\resizebox{1\linewidth}{!}
	{
	\setlength\tabcolsep{2pt}
	\begin{tabular}{l cccccccccc cccccccccc}

    \toprule
    Method&mIoU&car&bcycle.&mcycle.&truck&o-veh.&person&bclist.&mclist.&road&parking&side.&o-gro.&bui.&fence&vege.&trunk&terrain&pole&tra.\\
	\midrule
    KPConv&52.9&87.8&43.2&46.5&32.6&40.1&56.4&59.9&10.1&80.0&49.2&67.9&25.5&78.3&59.4&71.8&54.7&58.9&48.7&34.4\\ 
    KPConv w/ AziNorm&54.0&88.5&41.2&46.8&37.0&42.5&57.9&59.7&11.0&80.8&52.8&68.5&27.8&79.4&59.5&72.7&55.7&59.6&48.7&35.6\\ 
    Improvement&+1.1&+0.7&-2.0&+0.3&+0.8&+2.4&+1.5&-0.2&+0.9&+0.8&+3.6&+0.6&+2.3&+1.1&+0.1&+0.9&+1.0&+0.7&+0.0&+1.1\\
    \bottomrule
	\end{tabular}
	}
	\caption{\textbf{Performance comparisons of semantic segmentation on the SemanticKitti~\cite{SemanticKITTI} \textbf{test} set.} All experiments are conducted with the same configuration based on the official code of KPConv~\cite{KPConv}. AziNorm-based KPConv significantly outperforms vanilla KPConv.}
	\label{tab:semantickitti_test}
\end{table*}

\subsection{AziNorm-based Semantic Segmentation}
\noindent\textbf{Experimental Settings} \ 
For further validating the effectiveness and generalization ability, we integrate AziNorm into a widely-used semantic segmentation method named KPConv~\cite{KPConv}. We follow the released official code and only ablate on AziNorm. All hyper-parameters are kept the same for fair comparison.

\noindent\textbf{Dataset and Metrics} \ 
We conduct experiments on SemanticKitti~\cite{SemanticKITTI}, which
is a large-scale dataset based on the KITTI Vision Benchmark~\cite{Kitti} with dense point-wise annotations for the complete $360^\circ$ FOV. As official guidance~\cite{SemanticKITTI} suggests, we use mean intersection-over-union (mIoU) over all classes as the evaluation metric. 

\noindent\textbf{KPConv w/ AziNorm} \ 
Tab.~\ref{tab:semantickitti_val} and Tab.~\ref{tab:semantickitti_test} are the results of SemanticKitti. AziNorm-based KPConv significantly outperforms vanilla KPConv by $1.6\%$ and $1.1\%$ mIoU  on val  and test sets respectively. 
The experiments prove that AziNorm works in both object detection and semantic segmentation tasks. AziNorm is highly compatible and can also be applied in other LiDAR-based perception tasks.

\subsection{Data Efficiency \& Convergence}
\label{sec:convergence}
Apart from boosting perception performance, AziNorm can also significantly improve data efficiency and accelerates convergence.
In Fig.~\ref{fig:convergence}, under different data amounts and training epochs, we compare the performance (mAPH of LEVEL 2 difficulty for all classes) of SECOND detector w/ AziNorm and w/o AziNorm on Waymo. All other hyper-parameters are kept the same. With data amounts and training epochs reduced, the performance of SECOND drops dramatically, but SECOND w/ AziNorm  still achieves comparable performance.
SECOND w/ AziNorm  trained with only $3$ epochs outperforms vanilla SECOND  trained with $30$ epochs.
And SECOND w/ AziNorm  trained with only $10\%$ data outperforms vanilla SECOND trained with $100\%$ data.
AziNorm significantly improves data efficiency and accelerates convergence, reducing the requirement of both data amounts and training epochs by an order of magnitude.

From the perspective of machine learning, a perception method aims at approximating a function that maps from the point cloud patterns to the labels (\eg, 3D bounding boxes for detection and point-wise class labels for segmentation).
AziNorm treats the radial symmetry as an important inductive bias of the system.
It unifies symmetric patterns together and reduces the variability of input data.
With less variability in input, the mapping function is highly simplified and easier to be approximated.
Consequently, both the  data efficiency and convergence speed are significantly improved.

\subsection{Limitation}
\noindent\textbf{Overhead of AziNorm} \ 
AziNorm brings some computation overhead because of the overlap among patches.
But it is actually a problem of speed-accuracy trade-off. We can flexibly adjust the degree of overlap to balance between performance and inference speed.

For offline applications, \eg, constructing the high-definition map or generating 3D labels for camera-based systems (3D Data Auto Labeling), perception performance has the highest priority and latency is not restricted. We can make patches highly overlap to reach the performance upper bound of AziNorm.

For real-time applications, \eg, real-time perception in autopilot system,  it's feasible to reduce the overlap of patch for faster inference. AziNorm can still bring  gain even with little overlap. In Tab.~\ref{tab:speed}, we provide the speed comparison between SECOND and SECOND w/ AziNorm under the setting of little overlap ($r=11.2m$, $d=18.8m$, trained for $5$ epochs). SECOND w/ AziNorm achieves better performance and comparable inference speed.

\begin{table}
    \centering
    \resizebox{\linewidth}{!}{
		\begin{tabular}{l ccccc}
            \toprule
			Method & L1 mAP & L1 mAPH & L2 mAP & L mAPH & FPS\\
			\midrule
SECOND & 51.99 & 45.16 & 46.35 & 40.31 & 27\\
SECOND w/ AziNorm & 56.55 & 51.10 & 50.65 & 45.89 & 24\\
			\bottomrule
		\end{tabular}
	}
	\caption{\textbf{Inference speed comparison}. Tested on RTX 3090. Batch size is 1.  SECOND  w/  AziNorm  achieves better performance  and  comparable inference speed.}
	\label{tab:speed}
\end{table}

\subsection{Ablation Study}
\label{sec:abla}
In this section, we ablate on detailed elements of AziNorm on Waymo Open Dataset~\cite{waymo}, to validate the design of AziNorm. SECOND~\cite{second} detector is chosen for the experiments, given its compact and efficient pipeline.

\noindent\textbf{Patch Splitting, Translation Trans. and Rotation Trans.} \ 
In Tab.~\ref{tab:abla_normal}, we provide ablation studies about patch splitting, translation transformation and rotation transformation, to show what makes AziNorm work. The experiments are trained for $5$ epochs. 
The patch splitting mechanism can be considered as a kind of sampling strategy or data augmentation, and translation transformation narrows the numeric range of point coordinate and benefits convergence. Thus, patch splitting and translation transformation both bring gain. But their gain is limited, and the improvement of AziNorm mainly comes from rotation transformation.
Rotation transformation is the key step of AziNorm, which normalizes the patches along the radial direction.
It brings the most gain as expected, validating the motivation of this work, \ie, exploiting the radial symmetry of point cloud.

\begin{table}
    \centering
	\resizebox{1\linewidth}{!}
	{
	\begin{tabular}{l cccc }
    \toprule
    Method & L1 mAP &L1 mAPH & L2 mAP & L2 mAPH\\
	\midrule
    SECOND & 51.99 & 45.16 & 46.35 & 40.31\\
    \ \ + Patch Splitting & 53.07 & 46.87 & 47.52 & 42.04\\
    \ \ + Translation Trans. & 54.18 & 47.83 & 47.67 & 43.10\\
    \ \ + Rotation Trans. & 64.74 & 59.99 & 58.41 & 54.22\\
    \bottomrule
	\end{tabular}
	}
	\caption{\textbf{Ablation studies about patch splitting, translation transformation and rotation transformation.}  The improvement of AziNorm mainly comes from rotation transformation, which normalizes the point clouds along the radial direction.}
	\label{tab:abla_normal}
\end{table}

\noindent\textbf{Radius} $\boldsymbol{r}$ \noindent\textbf{and Stride} $\boldsymbol{d}$ \ 
Ablation studies about radius $r$ and stride $d$ are reported in Tab.~\ref{tab:abla_r} and Tab.~\ref{tab:abla_d} (trained for $5$ epochs). The performance of AziNorm is robust to radius $r$ and stride $d$ in a wide range. And when applied to different datasets, there is no need to specially tune $r$ and $d$.

\begin{table}
    \centering
	\resizebox{1\linewidth}{!}
	{
	\begin{tabular}{l ccccc }
    \toprule
    Method & $r$  & L1 mAP &L1 mAPH & L2 mAP & L2 mAPH\\
	\midrule
    SECOND & - & 51.99 & 45.16 & 46.35 & 40.31\\
    SECOND w/ AziNorm & 8.0m & 63.44 & 58.43 & 57.17 & 52.77\\
    SECOND w/ AziNorm & 9.6m & 64.74 & 59.99 & 58.41 & 54.22\\
    SECOND w/ AziNorm & 11.2m & 65.54 & 60.34 & 59.13 & 54.54\\
    \bottomrule
	\end{tabular}
	}
	\caption{\textbf{Ablation studies about radius} $\boldsymbol{r}$\textbf{.} Stride $d$ is set as $6.4m$. The performance of AziNorm is robust to radius $r$ in a wide range.}
	\label{tab:abla_r}
\end{table}

\begin{table}
    \centering
	\resizebox{1\linewidth}{!}
	{
	\begin{tabular}{l ccccc }
    \toprule
    Method  & $d$ & L1 mAP &L1 mAPH & L2 mAP & L2 mAPH\\
	\midrule
    SECOND & -  & 51.99 & 45.16 & 46.35 & 40.31\\
    SECOND w/ AziNorm & 8.0m & 63.56 & 58.71 & 57.53 & 53.15\\
    SECOND w/ AziNorm & 7.2m & 64.74 & 59.64 & 58.44 & 53.95\\
    SECOND w/ AziNorm & 6.4m & 64.74 & 59.99 & 58.41 & 54.22\\
    SECOND w/ AziNorm & 5.6m & 65.41 & 60.16 & 59.06 & 54.44\\
    \bottomrule
	\end{tabular}
	}
	\caption{\textbf{Ablation studies about radius} $\boldsymbol{d}$\textbf{.} Radius $r$ is set as $9.6m$. The performance of AziNorm is robust to stride $d$ in a wide range.}
	\label{tab:abla_d}
\end{table}

\noindent\textbf{Patch Layout} \ 
Ablation studies about the layout of patch are reported in Tab.~\ref{tab:abla_layout} (trained for $5$ epochs).
For fair comparison, we make the areas of these two kinds of patch similar. Circular patches and square patches have similar performance. Thus, AziNorm is robust to the choice of patch layout. We adopt circular patches for method description.

\begin{table}
    \centering
	\resizebox{\linewidth}{!}
	{
	\begin{tabular}{l ccccc }
    \toprule
    Patch layout & Area & L1 mAP &L1 mAPH & L2 mAP & L2 mAPH\\
    \midrule
    Circular & 290$m^2$ ($r$=9.6m) & 64.74 & 59.99 & 58.41 & 54.22\\
    Square & 310$m^2$ ($a$=17.6m) & 65.41 & 60.45 & 59.04 & 54.66\\
    \bottomrule
	\end{tabular}
	}
	\caption{\textbf{Ablation studies about the layout of patch.} Stride $d$ is $6.4m$.  "$a$" denotes the length of side. AziNorm is robust to the choice of patch layout.}
	\label{tab:abla_layout}
\end{table} 

\begin{figure}[]
    \centering
    \includegraphics[width=\linewidth]{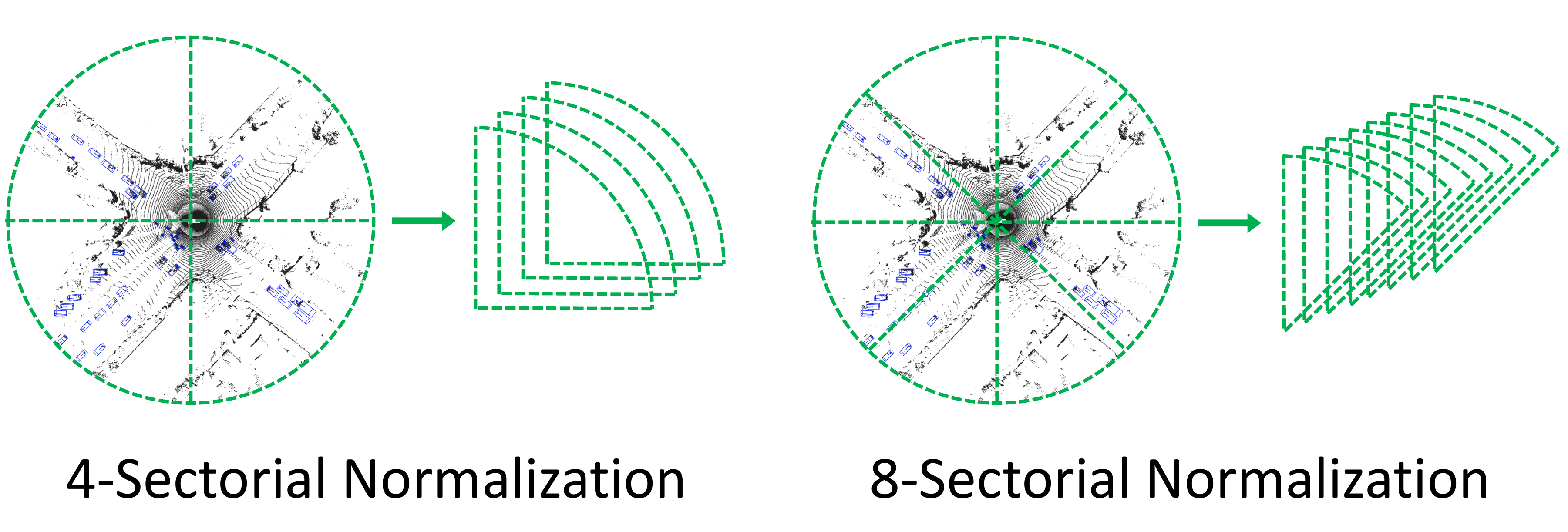}
    \caption{\textbf{$\boldsymbol{4}$- and $\boldsymbol{8}$-Sectorial Normalization.} The LiDAR scene is split into sectorial regions for normalization.}
    \label{fig:sectorial}
\end{figure}

\begin{table*}
    \centering
	\small
    {
	\begin{tabular}{c  l  cc  cc  cc  cc}
                \toprule
				\multirow{2}{*}{Difficulty} &
                \multirow{2}{*}{Method} & 
				\multicolumn{2}{c}{~~ All} &  
				\multicolumn{2}{c}{~~ Vehicle} &  
				\multicolumn{2}{c}{~~Pedestrian} & 
				\multicolumn{2}{c}{~Cyclist} \\
				& &mAP & mAPH & AP & APH & AP & APH & AP & APH \\
				\midrule
			    \multirow{4}{*}{LEVEL 1} 
    		    & w/o Norm. & 60.57 & 56.49 & 68.15 & 67.56 & 58.20 & 47.94 & 55.35 & 53.98\\
    		    & 4-Sectorial Norm. & 61.58 & 57.91 & 69.26 & 68.70 & 58.91 & 49.57 & 56.58 & 55.45\\
    		   	& 8-Sectorial Norm. & 62.21& 58.36 & 68.78 & 68.21 & 60.32 & 50.71 & 57.53 & 56.15 \\
    		    & AziNorm & \textbf{67.32} & \textbf{63.40} & \textbf{70.73} & \textbf{70.24} & \textbf{67.39} & \textbf{57.23} & \textbf{63.84} & \textbf{62.74}\\
    		    \midrule
			    \multirow{4}{*}{LEVEL 2} 
				& w/o Norm. & 54.48 & 50.90 & 59.61 & 59.09 & 50.22 & 41.32 & 53.62 & 52.28 \\
			    & 4-Sectorial Norm. & 55.66 & 52.37 & 60.66 & 60.16 & 51.50 & 43.24 & 54.82 & 53.72\\
			    & 8-Sectorial Norm. & 56.26 & 52.81 & 60.18 & 59.68 & 52.84 & 44.32 & 55.77 & 54.43\\
    			& AziNorm & \textbf{61.51} & \textbf{57.93} & \textbf{63.03} & \textbf{62.56} & \textbf{59.73} & \textbf{50.55} & \textbf{61.76} & \textbf{60.69} \\
				\bottomrule
			\end{tabular}   
			}
	\caption{\textbf{Experiments of Sectorial Normalization and AziNorm based on SECOND~\cite{second} on the Waymo~\cite{waymo} val set.} The comparisons between Sectorial Normalization and baseline validates the potential of the radial symmetry. And comparisons between AziNorm and 4-/8-Sectorial Normalization prove that finer normalization granularity brings more significant gain.}
	\label{tab:sectorial}
\end{table*} 

\begin{figure}[]
    \centering
    \includegraphics[width=\linewidth]{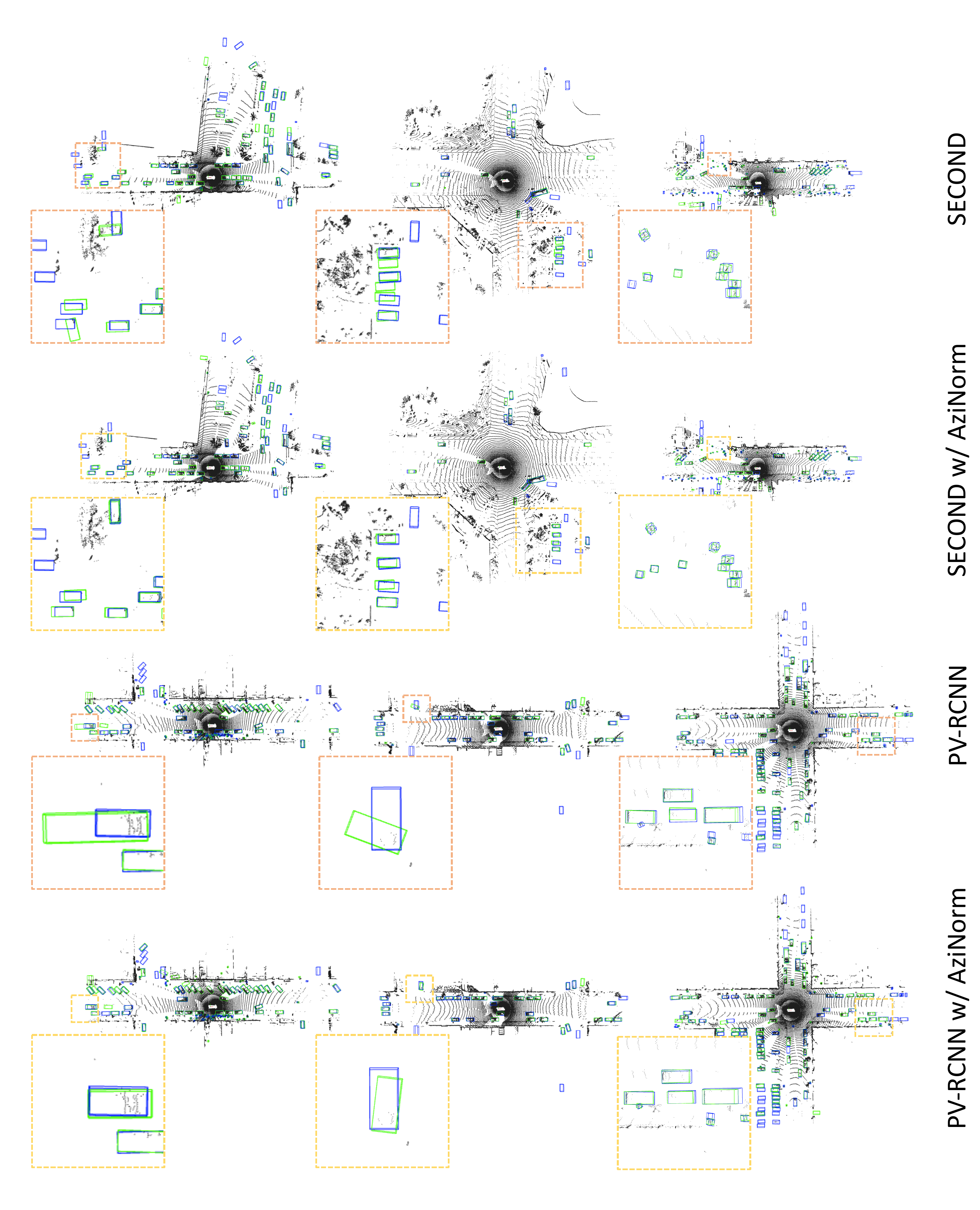}
    \caption{\textbf{Qualitative comparisons of AziNorm based on SECOND~\cite{second} and PV-RCNN~\cite{pvrcnn}.} Blue and green boxes are ground-truth annotations and predicted results respectively. When integrated with AziNorm, SECOND and PV-RCNN both achieve more accurate bounding box predictions, especially for the direction of bounding box. Key regions are zoomed in for better view.}
    \label{fig:visual}
\end{figure}

\noindent\textbf{Sectorial Normalization} \ 
\label{sec:norm_strategy}
Apart from AziNorm, we also design Sectorial Normalization (see Fig.~\ref{fig:sectorial}). The LiDAR scene is split into sectorial regions for normalization. Adjacent sectorial regions are slightly overlapped.
Tab.~\ref{tab:sectorial} shows the comparisons between Sectorial Normalization and AziNorm (based on SECOND and trained for $30$ epochs).  SECOND w/ $4$-Sectorial Normalization outperforms vanilla SECOND w/o normalization by $1.47$ mAPH of LEVEL 2 difficulty for all classes.  And SECOND w/ $8$-Sectorial Normalization achieves a higher improvement of $1.91$ mAPH.
AziNorm achieves the highest improvement of $7.03$ mAPH.

Compared with the baseline (w/o norm.), Sectorial Normalization changes little, no sampling strategy, no translation transformation and negligible computation overhead. It just splits the scene into sectorial regions and rotates them. With this simple change, we can get considerable gain. It proves the idea of exploiting the radial symmetry is effective and potential.

Besides, the experiments prove that finer \textbf{normalization granularity} brings more significant gain. $4$- and $8$-Sectorial Normalization reduce the azimuth variability to a range of $90^\circ$ and $45^\circ$ respectively. They help to simplify the point clouds but the normalization granularity is limited.
AziNorm corresponds to a finer normalization granularity, \ie, patch. The azimuth variability within a patch is negligible because of the limited spatial range of patch.
And the variability among patches is totally normalized through patch-specific transformation.
Thus, AziNorm almost eliminates the azimuth variability of the whole scene. It makes the most of the radial symmetry and achieves the most significant improvements.

\subsection{Qualitative Comparisons}
In Fig~\ref{fig:visual}, we provide qualitative comparisons
to further demonstrate the effectiveness of AziNorm.
When integrated with AziNorm, SECOND and PV-RCNN both achieve more accurate bounding box predictions, especially for the direction of bounding box, which is quite important for predicting the driver's intention in autopilot.

\section{Conclusion}
We present AziNorm, which exploits the inherent radial symmetry to normalize point clouds along the radial direction and eliminate the variability brought by the difference of azimuth. 
AziNorm is of great practical value, especially for autopilot and robotics. AziNorm 1) significantly boosts the perception performance, 2) improves data efficiency and lowers the cost of data acquisition and labeling, 3) accelerates convergence and saves training time.
Moreover, AziNorm is highly extensible. It can be easily combined with many perception tasks (detection, segmentation, etc.), and sensors (LiDAR, RADAR, surround-view cameras, etc.) that possess the same radial symmetry.

\paragraph{Acknowledgement} This work was in part supported by NSFC (No. 61733007 and No. 61876212) and the Zhejiang Laboratory under Grant 2019NB0AB02.

{\small
\bibliographystyle{ieee_fullname}
\bibliography{cvpr}
}

\end{document}